\documentclass[10pt, twocolumn, twoside, journal]{IEEEtran}

\usepackage{cite,amsmath}
\usepackage{amssymb, graphicx}
\usepackage{relsize} 
\usepackage{multicol}
\usepackage{multirow}
\usepackage{paralist}
\usepackage{url}

\usepackage{color}  

\begin{document}
\title{Tied Probabilistic Linear Discriminant Analysis \\ for Speech Recognition}
%
\author{Liang~Lu and Steve Renals
\thanks{Liang Lu, and Steve Renals are with University of Edinburgh, UK; email: {\smaller \tt \{liang.lu, s.renals\}@ed.ac.uk}}%
\thanks{The research was supported by EPSRC Programme Grant EP/I031022/1 (Natural Speech Technology).}
}

%
%
%
%
\maketitle
\begin{abstract}
Acoustic models using probabilistic linear discriminant analysis (PLDA) capture the correlations within feature vectors using subspaces which do not vastly expand the model. This allows high dimensional and correlated feature spaces to be used, without requiring the estimation of multiple high dimension covariance matrices. In this letter we extend the recently presented PLDA mixture model for speech recognition through a tied PLDA approach, which is better able to control the model size to avoid overfitting. We carried out experiments uisng the Switchboard corpus, with both mel frequency cepstral coefficient features and bottleneck feature derived from a deep neural network. Reductions in word error rate were obtained by using tied PLDA, compared with the PLDA mixture model, subspace Gaussian mixture models, and deep neural networks.
 
\end{abstract}


\section{Introduction}
\label{sec:intro}

\IEEEPARstart{A}{coustic} models for speech recognition have advanced substantially over the past 25 years, but the front-end feature processing has been largely unchanged, based on mel frequency cepstral coefficients (MFCCs) \cite{davis1980comparison} and perceptual linear prediction (PLP) features \cite{hermansky1990perceptual}.  To a large degree this has been due to the use of acoustic models based on hidden Markov models (HMMs) with Gaussian mixture models (GMMs) \cite{young1996review,gauvain2000large,gales2008application}, which are well matched to feature representations which have decorrelated components and are relatively low-dimensional. 

Deep neeural network (DNN) acoustic models \cite{dahl2012context} address these limitations and have achieved significant reductions in word error rate (WER) across many speech recogniiton datasets \cite{hinton2012deep}.  Compared to the hybrid neural network / hidden Markov model (HMM) architecture studied in the early 1990s \cite{bourlard1994connectionist, renals1994connectionist}, DNNs typically use more hidden layers and a wider output layer. Moreover, DNNs can be also used as a good feature extractor, for instance through the inference of bottleneck features which may append the features used in GMM-based speech recognition systems \cite{grezl2007probabilistic, yu2011improved}.  However, in order to be compatible with GMMs using diagonal covariances, such augmented feature vectors must typically be relatively low-dimensional and decorrelated.

We have addressed the limitations of GMMs through an acoustic model based on probabilistic linear discriminant analysis (PLDA) \cite{llu_spl14}, which can employ higher dimensional, correlated feature vectors. PLDA is a probabilistic extension of linear discriminant analysis (LDA) \cite{prince2007probabilistic}, which has been very well studied for speaker recognition in the joint factor analysis (JFA) \cite{kenny2007joint} and i-vector \cite{kenny2010bayesian, dehak2011front, matejka2011full} frameworks.  A PLDA acoustic model factorizes the acoustic variability using HMM state dependent variables which are expected to be consistent across different acoustic conditions, and observation dependent variables which characterise per frame level acoustic changes \cite{llu_spl14}. Similarly to a subspace GMM (SGMM) \cite{povey2010subspaceGMM}, the factorisation is based on the inference of subspaces.  However, while the SGMM uses a set of full covariance matrices to directly model the per frame acoustic variability, the PLDA model introduces another set of projections to model this variability in lower-dimension subspaces.    

We have previously investigated using a PLDA mixture model for acoustic modelling \cite{llu_spl14, llu_is2014}. Though good results have been obtained, this model has a large number of HMM state dependent variables, and is thus prone to overfitting.  In this letter we mitigate the problem by tying the PLDA state variables in PLDA, an approach analogous to the use of tied state vectors in SGMMs \cite{povey2010subspaceGMM}. 

\section{PLDA-based Acoustic Model}
\label{sec:plda}

The PLDA-based acoustic model is a generative model in which the distribution over acoustic feature vectors $\mathbf{y}_{t} \in \mathbb{R}^d$ from the $j$-th HMM state at time $t$ is expressed as:
\begin{align}
\mathbf{y}_{t} | j = \mathbf{U}\mathbf{x}_{jt} + \mathbf{G}\mathbf{z}_{j} + \mathbf{b} + \epsilon_{jt}, \quad \epsilon_{jt} \sim \mathcal{N}(\mathbf{0}, \Lambda) \, .
\end{align}
$\mathbf{z}_j \in \mathbb{R}^q$ is the state variable (equivalent to the between-class identity variable in JFA) shared by the whole set of acoustic frames generated by the $j$-th state and $\mathbf{x}_{jt} \in \mathbb{R}^p$ is the frame variable (equivalent to the within-class  channel variable in JFA) which explains the per-frame variability.  Usually, the dimensionality of these two latent variables is smaller than that of the feature vector $\mathbf{y}_t$, i.e. $p,q\le d$.  $\mathbf{U}\in \mathbb{R}^{d\times p}$ and $\mathbf{G} \in \mathbb{R}^{d\times q}$ are two low rank matrices which span the subspaces to capture the major variations for $\mathbf{x}_{jt}$ and $\mathbf{z}_j$ respectively. They are analogous to the within-class and between-class subspaces in the standard LDA formulation, but are estimated probabilistically. $\mathbf{b} \in \mathbb{R}^d$ denotes the bias and $\epsilon_{jt} \in \mathbb{R}^d$ is the residual noise which is assumed to be Gaussian with zero mean and diagonal covariance.  By marginalising out the residual noise variable $\epsilon_{jt}$, we obtain the following likelihood function:
\begin{align}
p(\mathbf{y}_t |\mathbf{x}_{jt}, \mathbf{z}_j, j) = \mathcal{N}(\mathbf{y}_t; \mathbf{U}\mathbf{x}_{jt} + \mathbf{G}\mathbf{z}_{j} + \mathbf{b}, \Lambda)
\end{align}

\subsection{PLDA Mixture Model}
\label{sec:mix}

A single PLDA has a limited modelling capacity since it only approximates a single Gaussian distribution.  An $M$-component PLDA mixture model \cite{llu_spl14}  results in the following component distribution:
\begin{align}
\label{eq:mix-plda}
\mathbf{y}_t | j, m &= \mathbf{U}_m\mathbf{x}_{jmt} + \mathbf{G}_m\mathbf{z}_{jm} + \mathbf{b}_m + \epsilon_{jmt}, \\
\epsilon_{jmt} &\sim \mathcal{N}(\mathbf{0}, \Lambda_m)
\end{align}
If $c$ to be the component indicator variable, then the prior (weight) of each component is $P(c=m |j) = \pi_{jm}$. Given the latent variables $\mathbf{x}_{jmt}$ and $\mathbf{z}_{jm}$, the state-level distribution over features is:
\begin{align*}
p(\mathbf{y}_t |j) = \sum_{m} \pi_{jm} \mathcal{N}(\mathbf{y}_t; \mathbf{U}_m\bar{\mathbf{x}}_{jmt} + \mathbf{G}_m\bar{\mathbf{z}}_{jm} + \mathbf{b}_m, \Lambda_m) \, .
\end{align*} 
$\bar{\mathbf{x}}_{jmt}$ and $\bar{\mathbf{z}}_{jm}$ are point estimates of the latent variables. 
Since the projection matrices $\mathbf{U}_m$ and $\mathbf{G}_m$ are globally shared,  a large number of components can be used to improve the model capacity, e.g. $M=400$ \cite{llu_spl14}. 

\subsection{Tied PLDA}
\label{sec:tied}
To avoid overfitting in the PLDA mixture model, those components which are responsible for a small number of feature vectors may be deactivated.  Alternatively, the state variables $\mathbf{z}_{jm}$ may be tied across components, resulting in the following component distribution:
\begin{align}  
 \label{eq:tied-plda}
\mathbf{y}_t | j, m &= \mathbf{U}_m\mathbf{x}_{jmt} + \mathbf{G}_m\mathbf{z}_{j} + \mathbf{b}_m + \epsilon_{jmt}, \\
\epsilon_{jmt} &\sim \mathcal{N}(\mathbf{0}, \Lambda_m) \, .
\end{align}
Tying the state variables may over-simplify the model. In this case, a ``mixing-up'' strategy can be used, analogous to SGMM sub-state splitting \cite{povey2010subspaceGMM}:
\begin{align}  
\label{eq:tied-plda2}
\mathbf{y}_t | j, k, m &= \mathbf{U}_m\mathbf{x}_{jkmt} + \mathbf{G}_m\mathbf{z}_{jk} + \mathbf{b}_m + \epsilon_{jkmt}, \\
\epsilon_{jkmt} &\sim \mathcal{N}(\mathbf{0}, \Lambda_m) \, ,
\end{align}
where
$k$ denotes the sub-state index, and $\mathbf{z}_{jk}$ is the sub-state variable. This makes Tied PLDA model more scalable compared to PLDA mixture model as we can balance the number of the sub-state variables according to the amount of available training data. Tied PLDA is equivalent to SGMM if we remove the per-frame latent variable $x_{jkmt}$ and use full covariance $\Lambda_m$ to model the residual noise. Given the latent variables, the state-level likelihood function can be written as
\begin{align}
\label{eq:tied-lik}
&p(\mathbf{y}_t | j)  = \sum_{mk}c_{jk}\times\pi_{jm} \mathcal{N}(\mathbf{y}_t; \mathbf{U}_m\bar{\mathbf{x}}_{jmkt} + \mathbf{G}_m\bar{\mathbf{z}}_{jk} + \mathbf{b}_m, \Lambda_m) \nonumber \\
& \quad = \sum_{mk}w_{jkm} \mathcal{N}(\mathbf{y}_t; \mathbf{U}_m\bar{\mathbf{x}}_{jmkt} + \mathbf{G}_m\bar{\mathbf{z}}_{jk} + \mathbf{b}_m, \Lambda_m)
\end{align} 
where $c_{jk}$ is the sub-state weight, $\pi_{jm}$ is the component weight which is shared for all the sub-state models, and $w_{jkm} = c_{jk}\times \pi_{jm}$. This is different to an SGMM in which a weight projection matrix is used to derive the component-dependent weights:
\begin{align}
\label{eq:sgmm}
p^{\mbox{\tiny SGMM}}(\mathbf{y}_t | j) &= \sum_{k}c_{jk}\sum_{m}\pi_{jkm} \mathcal{N}(\mathbf{y}_t;  \mathbf{G}_m\bar{\mathbf{z}}_{jk}, \pmb{\Sigma}_m)  \\
\label{eq:w-pro}
\pi^{\mbox{\tiny SGMM}}_{jkm} &= \frac{\exp \mathbf{w}_m^T\bar{\mathbf{z}}_{jk}}{\sum_{m'} \exp \mathbf{w}_{m'}^T\bar{\mathbf{z}}_{jk}}
\end{align}
where $\mathbf{w}$ denotes the weight projection matrix, and $\mathbf{w}_m$ denotes its $m$-th column. 
We do not use softmax weight normalisation in order to simplify the model training; empirical findings (Section \ref{sec:exp}) indicates that linear normalisation works well. Tied PLDA also differs from the SGMM by using another subspace projection (matrix $\mathbf{U}_m$) to model feature correlations. It is more scalable to high dimensional feature inputs than the direct feature covariance modelling used in SGMMs.

\section{Maximum Likelihood Training}
\label{sec:model-train}


\subsection{Likelihoods}
\label{sec:lik}


For tied PLDA, the likelihood may be computed according to equation (\ref{eq:tied-lik}) by make use of the MAP estimates of the latent variables $\mathbf{x}_{jkmt}$ and $\mathbf{z}_{jk}$, referred to as the  point estimate in \cite{llu_spl14}. However, this approach does not work well in practice because of the large uncertainty of the estimation of $\mathbf{x}_{jkmt}$, i.e. the large variance of its posterior distribution.

Another approach is to marginalise out the observation variable $\mathbf{x}_{jkmt}$, which is referred as the {\it uncertainty estimate} in \cite{llu_spl14}. Using $\mathcal{N}(\mathbf{0,I})$ as a prior, which is the same prior used in model training for consistency (cf. equation(\ref{eq:post_x})), this likelihood function can be obtained as
 \begin{align}
 \label{eq:unc}
 p(\mathbf{y}_t | j) &= \sum_{mk} w_{jkm} \int p(\mathbf{y}_t | \mathbf{x}_{jkmt}, j, k, m) P(\mathbf{x}_{jkmt})d\mathbf{x}_{jkmt} \nonumber \\
 &= \sum_{mk} w_{jkm} \mathcal{N}\left(\mathbf{y}_t; \mathbf{G}_m\bar{\mathbf{z}}_{jk} + \mathbf{b}_m, \mathbf{U}_m\mathbf{U}_m^T+\Lambda_m \right) \nonumber
 \end{align}    
This method is similar to the channel integration evaluation method used for JFA based speaker recognition \cite{glembek2009comparison, zhao2012variational}. Note that the likelihood can be efficiently computed without inverting matrices $\mathbf{U}_m\mathbf{U}_m^T+\Lambda_{m}$ directly, but by using the Woodbury matrix inversion lemma as in \cite{glembek2009comparison, li2012probabilistic}:
\begin{align}
&(\mathbf{U}_m\mathbf{U}_m^T+\Lambda_{m})^{-1} \nonumber \\
& \quad = \Lambda_m^{-1} - \Lambda_m^{-1}\mathbf{U}_m(\mathbf{I} + \mathbf{U}_m^T\Lambda_m^{-1}\mathbf{U}_m)^{-1}\mathbf{U}_m^T\Lambda_m^{-1} \\
& \quad = \Lambda_m^{-1} -  \mathbf{LL}^T
\end{align}
where $\mathbf{L} = \Lambda_m^{-1}\mathbf{U}_m(\mathbf{I}+\mathbf{U}_m^T\Lambda_m^{-1}\mathbf{U}_m)^{-1/2}$. This makes it computationally feasible when $\mathbf{y}_t$ is high dimensional. 

It is also possible to marginalise out the state variable $\mathbf{z}_{jk}$ alone or jointly with $\mathbf{x}_{jkmt}$ similar to the methods used in\cite{zhao2012variational}. However, we did not obtain a consistent improvement using this approach in our preliminary experiments. This may be the case because the variance of the posterior distribution of $\mathbf{z}_{jk}$ is small owing to increased training data  used for the posterior estimation. This model-based uncertainty approach is similar to Bayesian predictive classification (BPC) for GMM-based acoustic models \cite{huo2000bayesian}, in contrast to feature space uncertainty approaches used for noise robust speech recognition\cite{droppo2002uncertainty, liao2005joint, lu2013joint}.  
 
\subsection{Model update}
\label{model-update}
We used the Variational Bayesian inference to train the model where $\mathbf{x}_{jkmt}$ and $\mathbf{z}_{jk}$ are assumed  to be conditionally independent. A joint model training algorithm could be obtained without making use of this assumption:  however, it may be computationally infeasible in practice \cite{llu_is2014}.  Similar to the PLDA mixture model \cite{llu_spl14}, the EM auxiliary function to update $\mathbf{U}_m$ in tied PLDA is
\begin{align}
\label{aux_U}
\mathcal{Q}(\mathbf{U}_m) &= \sum_{jkt} \int P(j, k, m | \mathbf{y}_{t}) P(\mathbf{x}_{jkmt} | \mathbf{y}_{t}, \bar{\mathbf{z}}_{jk}, j, k, m)  \nonumber \\
&\quad \times \log p(\mathbf{y}_{t} | \mathbf{x}_{jkmt}, \bar{\mathbf{z}}_{jk}, j, k, m) d\mathbf{x}_t \nonumber \\
&= \sum_{jkt} \gamma_{jkmt}\mathbb{E} \Bigg[ -\frac{1}{2} \mathbf{x}_{jkmt}^T \mathbf{U}_m^T \Lambda_m^{-1} \mathbf{U}_m \mathbf{x}_{jkmt} \nonumber \\ 
&\quad + \mathbf{x}_{jkmt}^T \mathbf{U}_m^T \Lambda_m^{-1} \left(\mathbf{y}_{t} - \mathbf{G}_m \bar{\mathbf{z}}_{jk} - \mathbf{b}_m\right) \Bigg] + \text{const} \nonumber \\
&= \sum_{jkt} \gamma_{jkmt}\mathrm{Tr} \Bigg( \Lambda_m^{-1} \Big(-\frac{1}{2} \mathbf{U}_m \mathbb{E}[\mathbf{x}_{jkmt} \mathbf{x}_{jkmt}^T] \mathbf{U}_m^T \nonumber \\
& \quad + (\mathbf{y}_{t} - \mathbf{G}_m \bar{\mathbf{z}}_{jm} - \mathbf{b}_m) \mathbb{E}^T[\mathbf{x}_{jkmt}]\mathbf{U}_m^T \Big) \Bigg) + \text{const} \nonumber
\end{align}
where $\gamma_{jkmt}$ denotes the component posterior probability as
\begin{align}
\gamma_{jkmt} &= P(j, k, m | \mathbf{y}_t) \nonumber \\
&= P(j|\mathbf{y}_t)\frac{w_{jkm} p(\mathbf{y}_t | \bar{\mathbf{z}}_{jk}, j, k, m) }{\sum_{km}w_{jkm} p(\mathbf{y}_t | \bar{\mathbf{z}}_{jk}, j,k,m) } \, .
\end{align}
$P(j|\mathbf{y}_t)$ is the HMM state posterior which can be obtained using the forward-backward algorithm. $\mathbb{E}[\cdot]$ is the expectation operation over the posterior distribution of $\mathbf{x}_{jkmt}$:
\begin{align}
\label{eq:post_x}
&P(\mathbf{x}_{jkmt} | \mathbf{y}_{t}, \bar{\mathbf{z}}_{jk}, j, k, m) \nonumber \\
&\quad = \frac{p(\mathbf{y}_{t} | \mathbf{x}_{jkmt}, \bar{\mathbf{z}}_{jk}, j, k, m) P(\mathbf{x}_{jkmt})}{\int p(\mathbf{y}_{t} | \mathbf{x}_{jkmt}, \bar{\mathbf{z}}_{jk}, j, k, m) P(\mathbf{x}_{jkmt})d\mathbf{x}_{jkmt}}. 
\end{align}
Using $\mathcal{N}(\mathbf{0,I})$ as the prior distribution for $\mathbf{x}_{jkmt}$ we can obtain 
\begin{flalign}
& P(\mathbf{x}_{jkmt} | \mathbf{y}_{t}, \bar{\mathbf{z}}_{jk}, j, k, m)  = \mathcal{N}(\mathbf{x}_{jkmt}; \mathbf{V}_m^{-1}\mathbf{p}_{jkmt}, \mathbf{V}_m^{-1})\\
 & \mathbf{V}_m = \mathbf{I} + \mathbf{U}_m^{T}\Lambda_m^{-1}\mathbf{U}_m \\
 & \mathbf{p}_{jkmt} = \mathbf{U}_m^T\Lambda_m^{-1}(\mathbf{y}_{t} - \mathbf{G}_m \bar{\mathbf{z}}_{jk} - \mathbf{b}_m)
  \end{flalign}
 Note that using $\mathcal{N}(\mathbf{0,I})$ as a prior is reasonable since, after convergence, a nonzero mean can be accounted for by $\mathbf{b}_m$, and the variance can be modified by rotating and scaling the matrix $\mathbf{U}_m$. A similar form of posterior distribution can be obtained for $\mathbf{z}_{jk}$. 
 
By setting $\partial \mathcal{Q}(\mathbf{U}_m) / \partial \mathbf{U}_m = 0$ we obtain
\begin{align}
\mathbf{U}_m &= \left(\sum_{jkt} \gamma_{jkmt} (\mathbf{y}_{t} - \mathbf{G}_m\bar{\mathbf{z}}_{jk} - \mathbf{b}_m) \mathbb{E}^T[\mathbf{x}_{jkmt}] \right) \nonumber \\
& \quad \times \left( \sum_{jkt} \gamma_{jkmt} \mathbb{E}\left[\mathbf{x}_{jkmt} \mathbf{x}_{jkmt}^T\right] \right)^{-1}
\end{align}

Similarly, the update for other parameters are as follows.
\begin{align}
\mathbf{G}_m &= \left(\sum_{jkt} \gamma_{jkmt} (\mathbf{y}_{t} - \mathbf{U}_m\bar{\mathbf{x}}_{jkmt} - \mathbf{b}_m) \mathbb{E}^T[\mathbf{z}_{jk}] \right) \nonumber \\
& \quad \times \left( \sum_{jkt} \gamma_{jkmt} \mathbb{E}\left[\mathbf{z}_{jk} \mathbf{z}_{jk}^T\right] \right)^{-1}
\end{align}
\begin{align}
\mathbf{b}_m = \frac{\sum_{jkt}\gamma_{jkmt}(\mathbf{y}_t - \mathbf{U}_m\bar{\mathbf{x}}_{jkmt} - \mathbf{G}_m\bar{\mathbf{z}}_{jk})}{\sum_{jkt}\gamma_{jkmt}}
\end{align}
\begin{align}
\Lambda_{m}=\mbox{diag}\left( \frac{\sum_{jkt}\gamma_{jkmt}\left(\mathbf{y}_{jkmt}\mathbf{y}_{jkmt}^T + \mathbf{U}_m\mathbf{V}_m^{-1}\mathbf{U}_m^T\right)}{\sum_{jkt}\gamma_{jkmt}} \right)
\end{align}
where we have defined
\begin{align}
\mathbf{y}_{jkmt} = \mathbf{y}_t - \mathbf{U}_m\bar{\mathbf{x}}_{jkmt} - \mathbf{G}_m\bar{\mathbf{z}}_{jk} - \mathbf{b}_m
\end{align} 
The sub-state and component weights can be updated as
\begin{align}
c_{jk} = \frac{\sum_{mt}\gamma_{jkmt}}{\sum_{kmt}\gamma_{jkmt}}, \quad \pi_{jm} = \frac{\sum_{kt}\gamma_{jkmt}}{\sum_{kmt}\gamma_{jkmt}}
\end{align}
When using a large number of components, e.g. $M=400$ in this work, the weight should be floored by a small value for numerical stability. For computational efficiency, a background model based on a mixtures of factor analysers is used to select a small subset of the components for each frame for training and decoding, which is described in more detail in \cite{llu_spl14}. 

\begin{table*}[ht]
\caption{WER (\%) using 33 hours Switchboard training data, with different feature dimensions and different number of active model parameters} \vskip 1mm
\label{tab:baseline}
\centering \scriptsize
\begin{tabular}{l|lcccccc}
\hline \hline
System		& Feature  & Feature dim  & \#State-dependent parameters & \#State-indepdent parameters & CHM	& SWB		& Avg  \\ \hline
GMM & MFCC\_0+$\Delta$+$\Delta\Delta$ & 39 & $2.40\times 10^6$	 & - & 54.0	& 36.6	& 45.4 \\ 
GMM & MFCC\_0($\pm$2)+LDA\_STC & 40 & $2.43\times 10^6$ & - & 54.4 	& 34.4	& 43.7 \\ 
GMM & MFCC\_0($\pm$3)+LDA\_STC & 40 & $2.43 \times 10^6$ & - & 50.6 	& 33.5	& 42.2 \\ 
GMM & MFCC\_0($\pm$4)+LDA\_STC  & 40 & $2.43 \times 10^6$ & - & 50.7 	& 33.3	& 42.1 \\ 
GMM & MFCC\_0($\pm$5)+LDA\_STC  & 40 & $2.43 \times 10^6$ & - & 50.9 	& 34.1	& 42.4 \\ \hline
SGMM & MFCC\_0+$\Delta$+$\Delta\Delta$ & 39 & $0.8 \times 10^6$ & $0.97 \times 10^6$ &48.5 & 31.4   & 40.1 \\
SGMM & MFCC\_0($\pm$2)+LDA\_STC & 40 & $0.8 \times 10^6$ & $0.99 \times 10^6$& 45.7 & 30.0 & 38.0 \\ 
SGMM & MFCC\_0($\pm$3)+LDA\_STC & 40 & $0.8 \times 10^6$& $0.99 \times 10^6$ & 45.1 & 29.7 & 37.5 \\ 
SGMM & MFCC\_0($\pm$4)+LDA\_STC & 40 &$ 0.8 \times 10^6$ & $0.99 \times 10^6$ & 45.1 & 29.3 & 37.4 \\ 
SGMM & MFCC\_0($\pm$5)+LDA\_STC & 40 & $0.8 \times 10^6$ & $0.99 \times 10^6$& 45.7 & 29.5 & 37.7 \\ \hline
mix-PLDA & MFCC\_0 ($\pm$2) & 65 & $2.34 \times 10^6$ & $2.11 \times 10^6$ & 51.4 	& 33.1 	& 42.3 	\\
mix-PLDA & MFCC\_0 ($\pm$3) & 91 & $2.22 \times 10^6$ & $2.94 \times 10^6$ & 49.5 	& 32.4 	& 41.1 	\\
mix-PLDA & MFCC\_0 ($\pm$4)  & 117 & $2.16 \times 10^6$ & $3.78 \times 10^6$& 49.3	& 31.5	& 40.6 	\\ 
mix-PLDA & MFCC\_0 ($\pm$5) & 143 & $2.12 \times 10^6$ & $4.61 \times 10^6$ & 49.7	& 33.2	& 41.6   \\ \hline
tied-PLDA & MFCC\_0 ($\pm$2) & 65  & $0.86 \times 10^6$ & $2.11 \times 10^6$ & 48.6 	& 31.9 	& 40.4 	\\
tied-PLDA & MFCC\_0 ($\pm$3) & 91 & $0.86 \times 10^6$ & $2.94 \times 10^6$ & 47.9 	& 31.0 	& 39.5 	\\
tied-PLDA & MFCC\_0 ($\pm$4) & 117 &$0.86 \times 10^6$ & $3.78 \times 10^6$ & 47.5 	& 31.2 	& 39.4 	\\
tied-PLDA & MFCC\_0 ($\pm$5) & 143 & $0.86 \times 10^6$ & $4.61 \times 10^6$ & 48.7 	& 32.2 	& 40.6 	\\
tied-PLDA & MFCC\_0($\pm$3)+LDA\_STC & 40 & $0.85 \times 10^6$& $1.61 \times 10^6$ & 45.7 & 29.5 & 37.7 \\ \hline \hline
\end{tabular}
\end{table*}

\section{Experiments}
\label{sec:exp}
We performed experiments using the Switchboard corpus\footnote{\url{https://catalog.ldc.upenn.edu}} \cite{godfrey1992switchboard}. The Hub-5 Eval 2000 data \cite{cieri2002research} is used as the test set, which contains the Switchboard (SWB) and CallHome (CHM) evaluation subsets. The experiments were performed using the Kaldi speech recognition toolkit\footnote{\url{http://kaldi.sourceforge.net}} \cite{povey2011kaldi}, which we extended with an implementation of the PLDA-based acoustic model. In the following experiments, we have used maximum likelihood estimation without speaker adaptation or adaptive training. We used the pronunciation lexicon that was supplied by the Mississippi State transcriptions \cite{deshmukh1998resegmentation} and a trigram language model was used for decoding. 


\subsection{MFCC features}
\label{MFCC}

The first set of experiments used mel frequency cepstral coefficients (MFCCs) as features. We used the standard 39-dimensional MFCCs with first and second derivatives (MFCC$\_0\_\Delta\_\Delta\Delta$). To take advantage of longer context information, for the GMM and SGMM systems we have also performed experiments  using spliced MFCC\_0 of differing context window size, followed by a global LDA transformation to reduce the feature dimensionality to be 40, and a global semi-tied covariance (STC) matrix transform \cite{gales1999semi} to de-correlate the features. The PLDA systems directly used the concatenated MFCCs with various size of context window, without de-correlation and dimensionality reduction. 

\begin{table}[t]
\caption{WER (\%) using 33 and 109 hours Switchboard training data} \vskip 1mm
\label{tab:33-tandem}
\centering \footnotesize
\begin{tabular}{l | p{30mm}  p{4.5mm} p{4.5mm} p{4.5mm} p{4.5mm} }
\hline \hline
System	  & Feature &\multicolumn{2}{c}{33 hours} 	& \multicolumn{2}{c}{109 hours}  \\ 
& & CHM & SWB & CHM & SWB \\ \hline
DNN hybrid  & MFCC\_0+$\Delta$+$\Delta\Delta$ ($\pm$4) & 43.1 & 27.6 & 36.3 & 22.0 \\
BN hybrid & MFCC\_0+$\Delta$+$\Delta\Delta$ ($\pm$4) & 44.0 & 28.8 & 37.7 & 22.7 \\  \hline
GMM  & MFCC\_0+$\Delta$+$\Delta\Delta$  & 54.0	& 36.6	& 48.9 & 31.0 \\ 
GMM  & MFCC\_0($\pm$3)+LDA\_STC & 50.6 	& 33.5	& 44.9 & 28.0 \\ 
GMM  &BN\_MFCC &  44.8	& 30.9 & 39.7 & 25.5\\ 
GMM  & BN\_MFCC + LDA\_STC  & 43.2	& 27.4 & 36.7 & 22.1 \\ 
SGMM & BN\_MFCC + LDA\_STC & 41.7 & 26.7 & 36.2 & 21.7 \\ \hline 
mix-PLDA & BN\_MFCC & 42.6	& 27.1 & 35.9 & 21.6 \\ 
tied-PLDA & BN\_MFCC & 41.7	& 26.8 & 35.1 & 21.4 \\ \hline \hline
\end{tabular}
\end{table}

Table \ref{tab:baseline} shows the results of using a 33 hour subset of the training data, and the number of active model parameters\footnote{For PLDA systems, a component is considered active if its weight is above a threshold (0.01 in this work).}. In this case, there are about 2,400 clustered triphone states in the GMM systems, corresponding to about 30,000 Gaussians. The PLDA and SGMM systems have a similar number of clustered triphone states, and a 400-component background model is used for each. The state vector of SGMMs and latent variables of PLDA are all 40-dimensional. We used 20,000 sub-state vectors and state variables in the SGMM and tied PLDA systems, respectively. These results demonstrate the flexibility of PLDA systems in using different dimensional acoustic features, i.e. the spliced MFCC\_0 without any frontend feature transformations. Tied PLDA systems also offer consistently lower WERs than their counterparts based on the PLDA mixture model. Using the same low dimensional features as MFCC\_0($\pm$3)+LDA\_STC, the tied PLDA system achieved comparable recognition accuracy to SGMMs. This system is {\color{red} better than} tied PLDA systems using spliced MFCC\_0 of various context windows, which means that removing the non-discriminative dimensions in feature space is still beneficial to tied PLDAs. 

\subsection{Bottleneck features}
Table \ref{tab:33-tandem} shows the WERs of DNN and bottleneck systems using 33 hours and 109 hours of training data, respectively. The DNN system has six hidden layers, each with 1024 hidden units when using 33 hours of training data. The number of hidden units is increased to be 1200 when the amount of training data is 109 hours.The bottleneck DNN system (BN hybrid) used the same training data and the same kind of feature input --- while reducing the size of the fifth hidden layer to be 26. Using a larger bottleneck layer was not found to be helpful \cite{llu_is2014}. We concatenated the bottleneck and MFCC\_0+$\Delta$+$\Delta\Delta$ coefficients (referred as BN\_MFCC), and then used them to retrain our GMM and PLDA systems. We used LDA to reduce the dimensionality of the concatenated features from 65 to be 40 followed by STC to de-correlate the features for GMM and SGMM systems. Without the front-end feature transforms, the PLDA systems were able to achieve comparable or higher recognition accuracy by directly capturing the correlations between MFCCs and bottleneck features in subspaces.  Again, the results demonstrate the flexibility of PLDA acoustic models in terms of using input feature vectors of varying dimension.

\section{Conclusions}
\label{sec:sum}

Building upon our previous work on acoustic modelling using the PLDA mixture model, we have presented a tied PLDA based acoustic model, which is more scalable to the amount of training data. Experiments show that this model can achieve higher recognition accuracy while still enjoying the flexibility of using acoustic features of various dimension as the PLDA mixture model. Other types of acoustic feature representations can be more freely explored using this acoustic model. Along this line, we have demonstrated that the bottleneck feature from a DNN can used without any front-end feature transformation for dimensionality reduction and de-correlation. Future works include speaker adaptation and discriminative training for this model, and moreover, we are also interested in learning speech representations in an unsupervised fashion using a deep auto-encoder for this model. The source code and recipe used in this work are available from http://homepages.inf.ed.ac.uk/llu/code/plda-v1.tgz.  


\small
\bibliographystyle{IEEEbib}
\bibliography{bibtex}

\end{document}